\documentclass{article}
\usepackage{spconf,amsmath,graphicx, xcolor,amsfonts}

\title{Transformer-Based Sensor Fusion for Autonomous Driving: A Survey}
\name{Apoorv Singh\thanks{This literature review was done while working at Motional.}}
\address{Motional}
\begin{document}
%\ninept
%
\maketitle
\begin{abstract}
Sensor fusion is an essential topic in many perception systems, such as autonomous driving and robotics. Transformers-based detection head and CNN-based feature encoder to extract features from raw sensor-data has emerged as one of the top performing sensor-fusion 3D-detection-framework, according to the dataset leaderboards. In this work we provide an in-depth literature survey of transformer-based 3D-object detection task in the recent past, primarily focusing on the sensor fusion. We also briefly go through the Vision transformers (ViT) basics, so that readers can easily follow through the paper. Moreover, we also briefly go through few of the non-transformer based less-dominant methods for sensor fusion for autonomous driving. In conclusion we summarize the role that transformers play in the domain of sensor-fusion and also provoke future research in the field. More updated summary can be found at: https://github.com/ApoorvRoboticist/Transformers-Sensor-Fusion 

\end{abstract}
\section{Introduction}
\label{sec:intro}
Sensor fusion is a process of integrating sensory data from disparate information sources. With the complementary information captured by different sensors, fusion helps to reduce the uncertainty of state-estimation and make 3D object detection task more robust. Object attributes are not equally recognizable in different modalities, therefore we need to take advantage of different modalities and extract complementary information from them. For example LiDARs can do a better job in localizing a potential object; RADARs can do a better job in estimating velocity of the object in the scene; and last but not the least, cameras can do great job in classifying the object with its densely packed pixel information. \\ \\
\emph{What makes sensor-fusion problem so hard?} \\
Sensor data of different modalities usually has large discrepancies in the data-distribution; in addition to the difference in coordinate-space for each sensor. For example, natively, LiDAR  is in Cartesian coordinate-space; RADAR is in polar coordinate-space and images are in perspective-space. Spatial misalignment introduced by different coordinate frames makes it hard to merge these modalities together. Another issue with multi-modal input is that there would be asynchronous time-lines when camera and LiDAR feed is available to the ML network. 

\begin{figure}[htb]
  \centering
  \centerline{\includegraphics[width=8.5cm]{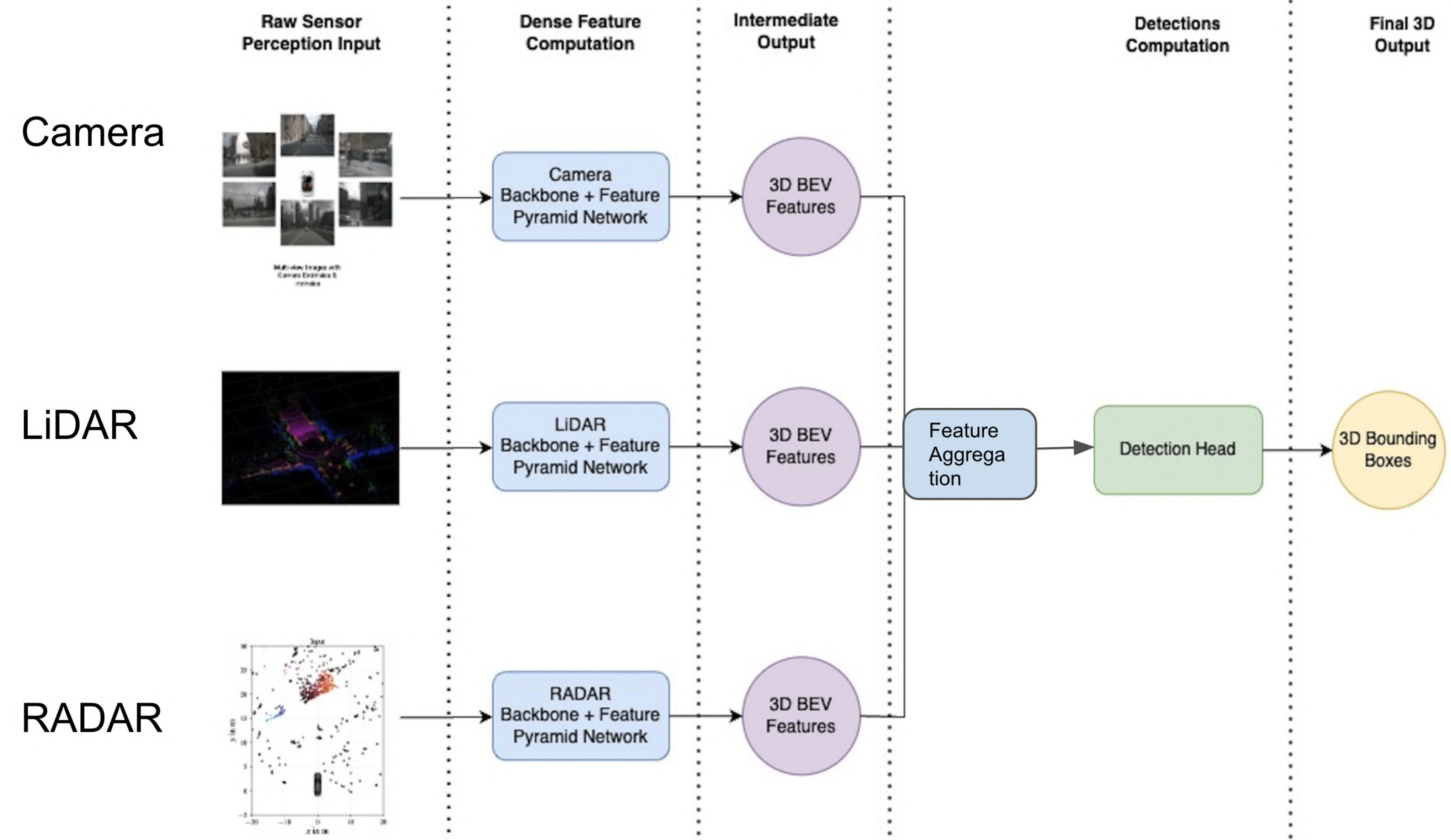}}
\caption{An overview architecture diagram of modern sensor-fusion model. Transformers-based head (Green-block); CNN-based Feature extractors (Blue-block) for predicting 3D Bird's Eye View (BEV) bounding box (Yellow-block) with intermediate BEV features per-sensor (Purple-block) as defined in Section \textcolor{blue}{\ref{sec:transformers_background}}. This sensor-fusion setup takes input from multi-view cameras, LiDARs and RADARs.}
\label{fig:sensor-fusion}
\end{figure}

While Deep-CNNs can be used to capture global context within a single modality, it is non-trivial to extend them to multiple modalities and accurately model interactions between pairs of features. To overcome this limitation, attention mechanisms of transformers is used to integrate global contextual reasoning about the 2D scene directly into the feature extraction layers of modalities. Recent advances in sequential modeling and audio-visual fusion \cite{video} demonstrates Transformers-based architecture is very competent in modeling the information interaction for sequential or cross-modal data. 

The main contributions of this work can be summarized as follows:
\begin{itemize}
    \item An overview for Vision Transformers(ViT) background to get readers up-to speed with the theoretical background prerequisite for going through the latest trending sensor fusion methods in Section \textcolor{blue}{\ref{sec:transformers_background}}
    \item Conduct an in-depth survey about the recent State-of-the-art(SoTA) methods for object detection task with sensor fusion, focusing on Transformers based approaches in Section \textcolor{blue}{\ref{sec:transformer-fusion}}
    \item Go through quantitative analysis of discussed SoTA methods and provoke future research work in the space in Section \textcolor{blue}{\ref{sec:quant_analysis}} and \textcolor{blue}{\ref{sec:conclusion}}.
\end{itemize}

\section{Related Work}
\label{sec:related_work}
\textbf{Fusion Levels:} Recently, multi-sensor fusion arouses increased interest in the 3D-detection community. Existing approach can be classified into \emph{detection-level}, \emph{proposal-level} and \emph{point-level} fusion methods, based on how early or late in the process we fuse different modalities viz., Cameras, RADARs, LiDARs et al.\\ \\\emph{Detection-level} a.k.a. late-fusion has emerged as the most simplest form of fusion, since each modality can process their own BEV detections individually which can be later post-processed to aggregate and remove duplicate-detections using Hungarian cost-matching algorithm and Kalman-filtering. However, this approach is not able to leverage the fact that each sensor can also contribute on different attributes within a single bounding-box prediction. CLOCS \cite{clocs} leverages modalities in the form they can naturally perform detection task i.e. LiDAR for 3D object detection and cameras for 2D detection task. It operates on the both the output candidates before Non-maximum suppression and use geometric consistencies between two set of predictions to get rid of False-positives (FP), as it is highly unlikely that the same FP would be detected simultaneously with different modalities. \\ \\
\emph{Point-level} a.k.a. early-fusion is augmenting LiDAR point-cloud with the camera features \cite{pointpainting, point-fusion}. In this method we find hard associations between LiDAR points and images using transformation matrices. However camera-to-LiDAR projections are semantically lossy as fusion quality is limited by the points sparsity. This approach suffers when there is even a slight error in calibration parameters of the two sensors. \\ \\ 
\emph{Proposal-level} a.k.a. deep-fusion is the most researched-upon method in the literature these days. Advances in transformers \cite{detr, detr3d, bevformer} have unlocked the possibilities of how intermediate features can interact despite being cross-domain from different sensors. Representative work like MV3D \cite{mv3d} proposes initial bounding boxes from LiDAR features and iteratively refines them using camera features. BEVFusion \cite{bevfusion} generates camera-based BEV features as highlighted in \cite{lss, singh20233m3d, singh2023surround, singh2023vision}. Camera and LiDAR modality are concatenated in the BEV space and a BEV decoder \cite{centerpoint} is used to predict 3D boxes as a final output. In TransFuser \cite{transfuser}, single-view image and LiDAR's BEV representation is fused by transformers in the encoder at various intermediate feature maps. This results in a 512-dimensional feature vector output of an encoder that constitutes a compact representation of the local and global context. In addition, this paper feeds the ouput to a GRU (Gated Recurrent Unit) and predict differentiable ego-vehicle way-points using L1 regression loss. 4D-Net \cite{4dnet}, in-addition to being multi-modal, adds temporal dimension to the problem as the $4^{th}$ dimension. They first extract in-time features of cameras and LiDAR\cite{pointpillars} individually first. To add different context of image representation; they collect image features in three-levels viz., high-resolution image, low-resolution image, video. Then they fuse cross-modal information using transformation matrix to fetch 2D context given 3D-pillar center, defined by the center-point of the BEV grid cell $(x^o, y^o, z^o)$. 
\section{Transformers Based Fusion Network Background}
\label{sec:transformers_background}
This approach can be divided into 3-steps: 1. Apply CNN-based backbones to extract spatial features from all the modalities individually. 2. Small set of learned embeddings (object queries/ proposals) are iteratively refined in transformers module to generate a set of predictions of 3D boxes. 3. Set-based loss is calculated over the predictions and groud-truth. This architecture is presented in Fig. \textcolor{blue}{\ref{fig:sensor-fusion}}
\subsection{Backbone: Feature Extractor}
\textbf{Cameras:} Multi-camera images are fed into the backbone network (e.g., ResNet-101) and FPN \cite{fpn} and obtain features $\{\{F^{ij}_{images}\in \mathbb{R}^{C*h*w}\}^{N_{view}}_{i=1}\}^M_{j=1}$, where M is the number of feature levels from FPN; $N_{view}$ is the number of cameras in surround-view; $h*w$ is the image-view feature dimensions.   
\\
\textbf{LiDAR:} Generally a voxelNet\cite{voxelnet} with 0.1m voxel size or PointPillar\cite{pointpillars} with 0.2m pillar size is used to encode points. After 3D backbone and FPN \cite{fpn}, a multi-scale BEV feature maps $\{F^j_{lidar}\in \mathbb{R}^{C*H_j*W_j}\}^M_{j=1}$ is obtained. \\
\textbf{RADARS:} Let's consider $N$ Radar points $\{r_j\}^N_{j=1} \in \mathbb{R}^{C_{radar}}$, where $C_{radar}$ is the number of features of the radar points, such as location, intensity and speed. A shared MLP $\Phi_{radar}$ is used to obtain per-point features $F^i_{rad}=\Phi_{rad}(r_j)\in\mathbb{R}^C$.

\subsection{Query Initialization}
In seminal work \cite{detr}, sparse queries are learned as a network parameter and is a representative of the entire training data. This type of query takes longer i.e. more number of sequential decoder layers (typically qty. 6) to iteratively converge to the actual 3d-objects in the scene. However, recently input-dependent queries\cite{efficientdetr} are considered as a better initialization strategy. This strategy can bring a 6-layered transformer decoder down to even a single-layered decoder layer. Transfusion \cite{transfusion} uses center-heatmap as queries and BEVFormer \cite{bevformer} introduced dense-queries as equally-spaced BEV grid.

\subsection{Transformers Decoder}
\begin{table*}[t]
\caption{Performance comparison on the nuScenes \textbf{test set}. Metrics key is defined in Section \textcolor{blue}{\ref{sec:quant_analysis}}.}
% \vspace{-2.0mm}
\begin{center}
\resizebox{2.0\columnwidth}{!}{
\begin{tabular}{|l|r|r|r|r|r|r|r|r|}
\hline
Methods & NDS(\%) $\uparrow$ & mAP(\%) $\uparrow$ & mATE(cm) $\downarrow$ & mASE(\%) $\downarrow$ & mAOE(rad) $\downarrow$ & mAVE(cm/s) $\downarrow$ & mAAE(\%) $\downarrow$ \\
\hline
CMT \cite{cmt} & 73.0 & 70.4 & 29.9 & 24.1 & 32.3 & 24.0 & 11.2 \\
BEVFusion \cite{bevfusion} & 72.9 & 70.2 & 26.1 & 23.9 & 32.9 & 26.0 & 13.4 \\
TransFusion \cite{transfusion} & 71.7 & 68.9 & 25.9 & 24.3 & 35.9 & 28.8 & 12.7 \\
UVTR \cite{uvtr} & 71.1 & 67.1 & 30.6 & 24.5 & 35.1 & 22.5 & 12.4 \\
\hline
\end{tabular}
}
\label{table:result}
\end{center}
\end{table*}
To refine object-proposals, repeated blocks of Transformer decoders are used sequentially in a ViT model, where each block consists of self-attention and cross-attention layers. \emph{Self-attention} between object queries does pairwise reasoning between different object candidates. \emph{Cross-attention} between the object queries and the feature-map aggregates relevant context into the object queries based on learned attention mechanism. Cross-attention is the slowest step in the chain because of the huge feature size, but techniques \cite{deformable} had been proposed to reduce the attention window. After these sequential decoders, d-dimensional refined queries are independently decoded with an FFN layers following \cite{centerpoint}. FFN predicts the center-offset from the query position $\delta{x},\delta{y}$, bounding box height as $z$, dimensions $l, w, h$ as $log(l), log(w), log(h)$, yaw angle $\alpha$ as $\sin{(\alpha)}$ and $\cos{(\alpha)}$ and velocity as $v_x, v_y$; lastly per-class probability $\hat{p}\in[0,1]^K$ is predicted for K semantic classes.

\subsection{Loss Computations}
Bipartite matching between set-based predictions and ground-truths through Hungarian algorithm is used, where matching cost is defined by: 
\begin{equation}
C_{match}=\lambda_1L_{cls} + \lambda_2L_{reg} + \lambda_2L_{IoU}
\end{equation}
where, $L_{cls}$ is a binary cross-entropy loss; $L_{reg}$ is an L1 loss; $L_{IoU}$ is a box IoU loss. $\lambda_1, \lambda_2, \lambda_3$ are network hyper-parameters.

\section{Transformers-based Sensor Fusion}
\label{sec:transformer-fusion}
\textbf{TransFusion} \cite{transfusion} tackles modality misalignment issue with the soft-association of features. First decoder layer constitutes sparse-queries generation from LiDAR BEV features. Second decoder layer enriches LiDAR queries with the image features with soft associations by leveraging locality inductive bias with cross-attention only around the bounding box decoded from the query. They also have image-guided query initialization layer. \\
\textbf{FUTR3D} \cite{futr3d} is closely related to \cite{detr3d}. It is robust to any number of sensor modalities. MAFS (Modality Agnostic Feature sampler) takes in the 3D queries and aggregates features from multi-view cameras, high-res lidars, low-res lidars and radars. To be specific, it first decodes query to get 3D-coordinate, which is then used as an anchor to gather features from all the modalities iteratively. BEV features are used for LiDAR and cameras, however for RADARS, top-k nearest radar points are picked in MAFS. For each query $i$, all these features $F$ are concatenated as below where $\Phi$ is an MLP layer:
\begin{equation}
    F^i_{fused} = \Phi_{fused}(F^i_{lidar} \oplus F^i_{camera} \oplus F^i_{radar})
\end{equation}
\textbf{CMT: Cross-Modal Transformers} \cite{cmt} encodes 3D coordinates into the multi-modal tokens by the \emph{coordinates encoding}. The queries from the \emph{position-guided query generator} are used to interact with the multi-modal tokens in transformer decoder and then predict the object parameters. \emph{Point-based query denoising} is further introduced to accelerate the training convergence by introducing local prior. \\
\textbf{UVTR: Unifying Voxel-based Representation with Transformer } \cite{uvtr} unifies multi-modality representations in the voxel-space for accurate and robust single or cross-modality 3D detection. Modality-specific space is first designed to represent different input in the voxel space without height compression to alleviate semantic ambiguity and enable spatial connections. This is a more complex and more information packed representation compared to the other BEV approaches. For \emph{image-voxel space}, perspective view features are transformed into the predefined space with view-transform, following \cite{lss}. CNN-based voxel encoder is introduced for multi-view feature interactions. For \emph{point-voxel space}, 3D-points can be naturally transformed into voxels. Sparse convolutions are used over these voxel features to aggregate spatial information.  With accurate positions in the point-cloud, the semantic ambiguity in $z$ direction is much reduced compared to the images.\\
\textbf{LIFT: LiDAR Image Fustin Transformer} \cite{lift} is capable to align the 4D spatiotemporal cross-sensor information. In contrast to \cite{4dnet} it exploits integrated utilization of sequential multi-modal data. For sequential data processing, they use the prior of vehicle pose to remove the effects of ego-motion between temporal data. They encode  both the lidar frames and camera images as sparsely located BEV grid features and proposes a sensor-time 4D attention module to capture mutual correlation.\\
\textbf{DeepInteraction:} \cite{deepinteraction}, follows a little different approach compare to its other counterparts. It claims that the previous approaches are structurally restricted due to its intrinsic limitations of potentially dropping off a large fraction of modality-specific representational strengths due to largely imperfect information fusion into a unified representation as in \cite{pointpainting, bevfusion}. They instead of deriving a fused single-BEV representation, they learn and maintain two modality-specific representations throughout to enable inter-modality interaction so that both information exchange and modality specific strengths can be achieved spontaneously. They refer to it as a multi-input-multi-output (MIMO) structure, taking as input of two modality-specific scene representations which are independently extracted by LiDAR and image backbones, and producing two refined representations as output. This paper includes DETR3D-like \cite{detr3d} queries which are sequentially updated from LiDAR and vision features with sequential cross-attention layers in the transformer-based decoder layer.\\
\textbf{Auto-align} \cite{autoalign} they model a mapping relationship between image and point-cloud with a learn-able alignment map, instead of establishing a deterministic correspondences for sensor projection matrix as done in the other approaches. This map enables the model to automate the alignment of non-homogeneous features in a dynamic data-driven manner. They leverage cross-attention module to adaptively aggregate pixel-level image features for each voxel.

\section{Quantitative Analysis}
\label{sec:quant_analysis}
Here we compare previously discussed methods on nuScenes \cite{nuscenes}, a large-scale multi-modal dataset, which is composed of data from 6 cameras, 1 LiDAR and 5 RADARs in Table \textcolor{blue}{\ref{table:result}}. This dataset has 1000 scenes total and is divided into $700/150/150$ scenes as train/validation/test sets, respectively. \textbf{Cameras:} Each scene has $20s$ video frame with 12 FPS. 3D bounding boxes are annotated at $0.5s$. Each sample includes 6 cameras. \textbf{LiDAR:} A 32-beam LiDAR with 20FPS is also annotated at every $0.5s$. \textbf{Metrics}: We follow the nuScenes official metrics. Keys are as follows: nuScenes Detection Score (NDS), mean Average Precision (mAP), mean Average Translation Error (mATE), mean Average Scale Error (mASE), mean Average Orientation Error(mAOE), mean Average Velocity Error (mAVE) and mean Average Attribute Error (mAAE).

\section{Conclusion}
\label{sec:conclusion}
For the autonomous vehicle's perception reliability, accurate 3D-object detection is one of the key challenge which we need to solve. Sensor-fusion helps making these predictions more accurate by leveraging pros of all the sensors present on the platform. Transformers have emerged as one of the top method to model these cross-modal interactions, especially when sensors operate in different coordinate space which makes it impossible to align perfectly. 

% \vfill\pagebreak
\bibliographystyle{IEEEbib}
\bibliography{refs}

\end{document}